\documentclass[lettersize,journal]{IEEEtran}
\usepackage{amsmath,amsfonts}
\usepackage{algorithmic}
\usepackage{algorithm}
\usepackage{array}
\usepackage[caption=false,font=normalsize,labelfont=sf,textfont=sf]{subfig}
\usepackage{adjustbox}  
\usepackage{multirow}   
\usepackage{graphicx}      
\usepackage{adjustbox}     
\usepackage{amsmath}       
\usepackage{amssymb}       
\usepackage{graphicx}
\usepackage{bigstrut}
\usepackage{textcomp}
\usepackage{stfloats}
\usepackage{url}
\usepackage{verbatim}
\usepackage{graphicx}
\usepackage{cite}
\usepackage{makecell}
\usepackage[colorlinks,
            urlcolor=blue,
            linkcolor=blue,
            anchorcolor=blue,
            ]{hyperref}
\begin{document}

\title{
A3-TTA: Adaptive Anchor Alignment Test-Time Adaptation for Image Segmentation}
\author{Jianghao Wu, Xiangde Luo, Yubo Zhou, Lianming Wu, Guotai Wang, Shaoting~Zhang
\thanks{This work was supported in part by the National Natural Science Foundation of China (62271115), and in part by the Natural Science Foundation of Sichuan Province (2025ZNSFSC0455).}
\thanks{Jianghao Wu, Xiangde Luo, Yubo Zhou, Guotai Wang and Shaoting Zhang are with the School of Mechanical and Electrical Engineering, University of Electronic Science and Technology of China, Chengdu, 611731, China. Guotai Wang and Shaoting Zhang are also with Shanghai AI laboratory, Shanghai, 200030, China. 
(e-mail: guotai.wang@uestc.edu.cn)}
\thanks{Lianming Wu is with Department of Radiology, Renji Hospital, School of Medicine, Shanghai Jiao Tong University, Shanghai, China. }
}

\maketitle
\markboth{IEEE TRANSACTIONS ON IMAGE PROCESSING, VOL. XX, NO. XX, XXXX 2025}{Jianghao Wu \MakeLowercase{\textit{et al.}}: A3-TTA: Adaptive Anchor Alignment Test-Time Adaptation}

\begin{abstract}
Test-Time Adaptation (TTA) offers a practical solution for deploying image segmentation models under domain shift without accessing source data or retraining. Among existing TTA strategies, pseudo-label-based methods have shown promising performance. However, they often rely on perturbation-ensemble heuristics (e.g., dropout sampling, test-time augmentation, Gaussian noise), which lack distributional grounding and yield unstable training signals. This can trigger error accumulation and catastrophic forgetting during adaptation.
To address this, we propose \textbf{A3-TTA}, a TTA framework that constructs reliable pseudo-labels through anchor-guided supervision. Specifically, we identify well-predicted target domain images using a class compact density metric, under the assumption that confident predictions imply distributional proximity to the source domain. These anchors serve as stable references to guide pseudo-label generation, which is further regularized via semantic consistency and boundary-aware entropy minimization. Additionally, we introduce a self-adaptive exponential moving average strategy to mitigate label noise and stabilize model update during adaptation.
{Evaluated on both multi-domain medical images (heart structure and prostate segmentation) and natural images}, A3-TTA significantly improves average Dice scores by 10.40 to 17.68 percentage points compared to the source model, outperforming several state-of-the-art TTA methods {under different segmentation model architectures}. 
A3-TTA also excels in continual TTA, maintaining high performance across sequential target domains with strong anti-forgetting ability.
The code will be made publicly available at \href{https://github.com/HiLab-git/A3-TTA}{https://github.com/HiLab-git/A3-TTA}.
\end{abstract}

\begin{IEEEkeywords}
Medical image segmentation, Test-time adaptation, Feature bank
\end{IEEEkeywords}

\section{Introduction}
\label{sec:introduction}

\begin{figure*}[t]
    \centering
\centerline{\includegraphics[width=18cm]{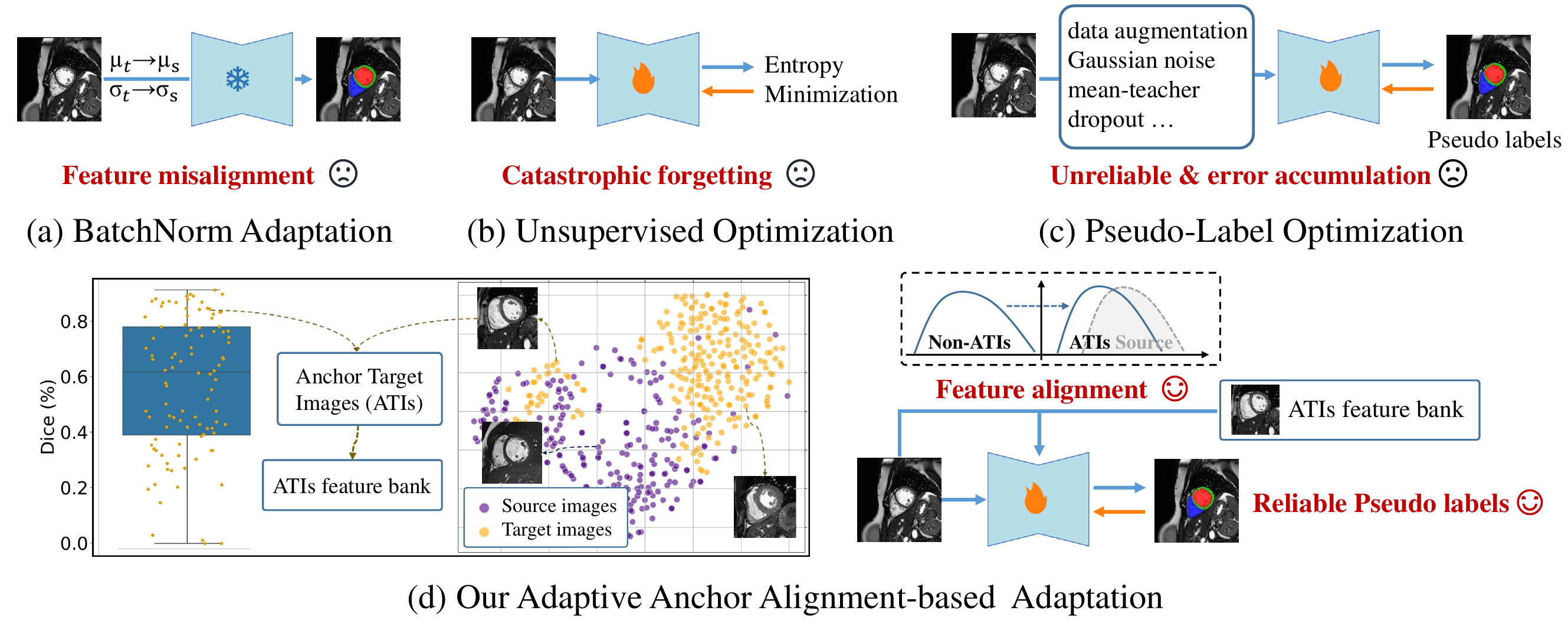}}
\caption{Motivation of our A3-TTA. (a) BatchNorm Adaptation suffers from feature misalignment due to mismatched statistics between source and target domains; (b) Unsupervised Optimization, such as entropy minimization, may lead to catastrophic forgetting of source knowledge; (c) Pseudo-Label Optimization is prone to unreliable pseudo labels and error accumulation; (d) Our Adaptive Anchor Alignment-based Adaptation (A3-TTA) mitigates these issues by leveraging Anchor Target Images (ATIs) to establish a feature bank, facilitating more stable feature alignment and generating reliable pseudo labels for adaptation.}
    \label{fig:fig1_anchor}
\end{figure*}

\IEEEPARstart{D}{eep} learning has become the dominant paradigm for {image segmentation}, delivering strong performance when the training and testing distributions are well aligned~\cite{litjens2017survey,zhu2023brain,liu2022sf}.  
However, domain shifts, {ranging from variations in imaging protocols, scanners, and patient populations in medical settings to weather, illumination, and scene dynamics in real-world environments}, often degrade performance in deployment~\cite{payette2023fetal,sun2021multi,xu2023novel,yue2022toward}.
To deal with this problem, several types of domain adaptation methods including Unsupervised Domain Adaptation (UDA)~\cite{dorent2023crossmoda,zheng2024dual} and Source-Free Unsupervised Domain Adaptation (SFUDA)~\cite{yang2022source,bateson2022source,jing2024visually} have been proposed to adapt models trained with a source domain to a new target domain. {UDA methods often assume the access of a set of labeled source domain images in the target domain, and use source- and target-domain images simultaneous for model training~\cite{wu2024fpl-plus}.  SFUDA avoids the use of source data, and only use a set of unlabeled target-domain data for model adaptation~\cite{yang2022source}. However, it still needs multi-epoch training in the target domain, which limits its feasibility in real-time clinical settings where data arrive in streams, which makes both availability of a large set of test-domain images and multi-epoch update not allowed. 
In contrast, fully Test-Time Adaptation (TTA)~\cite{wang2021tent} adapts models online using only unlabeled test images, without altering the original model or requiring source data, making it well-suited for deployment in dynamic clinical environments.}

Existing TTA methods can be broadly grouped into three categories: Batch Normalization (BN)-based, unsupervised optimization-based, and pseudo-label-based.  
BN-based methods~\cite{nado2020evaluating,dong2024medical} update or optimize BN statistics using target domain data, but suffer under large domain shifts due to mismatched feature distributions and the lack of pixel-level supervision.  
Unsupervised optimization-based methods, such as TENT~\cite{wang2021tent}, adapt the model by minimizing prediction entropy on unlabeled target data. While conceptually simple, these approaches provide task-agnostic and structure-unaware supervision, which often leads to unstable adaptation behavior in single-pass settings.
To enhance supervision, pseudo-label-based methods~\cite{wu2023upltta,wang2022continual,zhu2024improving} generate self-labels through perturbation-ensemble strategies, such as data augmentation~\cite{khurana2021sita}, Monte Carlo dropout~\cite{wu2023upl}, Gaussian noise injection~\cite{zhu2024improving}, and mean-teacher frameworks~\cite{wang2022continual}.  
While these methods improve label diversity, they primarily rely on stochastic perturbations of the model or input and lack explicit mechanisms to align with the source distribution.   
As a result, the generated pseudo-labels are often noisy and unstable, limiting their effectiveness as reliable supervision signals. 
Consequently, many existing TTA methods remain vulnerable to error accumulation or performance degradation, as illustrated in Fig.~\ref{fig:fig1_anchor}(a–c).

{Motivated by challenges of TTA in  segmentation tasks including unavailable source domain images, online adaptation and and boundary sensitivity, we propose \emph{A3-TTA}, an anchor-guided test-time adaptation framework that (i) replaces perturbation voting with ATI-driven bottleneck-level alignment, (ii) targets boundary ambiguity with boundary-aware entropy, and (iii) stabilizes non-stationary updates via a discrepancy-adaptive EMA teacher.}
As shown in Fig.~\ref{fig:fig1_anchor}(d), the source model exhibits markedly different performance across target images; a subset shows high softmax compactness and relatively accurate predictions. {We term these reliable, source-like target cases \emph{Anchor–Target Images (ATIs)} and leverage them on-the-fly to stabilize subsequent cases without any source data.}
To identify such samples, we introduce a Class Compactness Density metric based on softmax probability distributions and construct a fixed-size feature bank to store the representations of the selected ATIs. This feature bank serves as an intermediate-domain reference during adaptation.
For each incoming test image, we retrieve the most similar feature from the ATI bank and perform a feature alignment procedure through simple fusion and normalization operations, which help generate more reliable pseudo labels. These pseudo labels are subsequently used to supervise the model through a combination of semantic consistency enforcement and boundary-aware entropy minimization, improving pixel-level prediction quality.  
To further enhance stability during adaptation, we integrate a self-adaptive Exponential Moving Average (EMA) update strategy within a mean-teacher framework. The EMA rate is dynamically regulated by the normalized cross-entropy divergence between teacher and student predictions, allowing the teacher to update more aggressively when domain shifts are large and to preserve stability when predictions are consistent.
A3-TTA is model-agnostic, requires no source data or additional proxy tasks, and can be seamlessly applied to existing segmentation models without architectural modifications. It is designed to support real-time deployment under sequential and unlabeled testing scenarios.
Our contributions are summarized as follows:
\begin{itemize}
\item We introduce the concept of Anchor-Target Images (ATIs) and propose a Class Compactness Density metric to identify them based on softmax confidence.
\item We construct a dynamic feature bank of ATIs and perform feature alignment through fusion and normalization to refine the representations of incoming test images for pseudo-label generation.
\item We enhance adaptation via semantic consistency and boundary-aware entropy minimization, improving pixel-level supervision.
\item We introduce a self-adaptive EMA update that leverages teacher–student cross-entropy divergence to balance rapid adaptation and stability in single-pass test-time learning.
\end{itemize}

{We conducted experiments on both medical and natural images. For medical images, results show that A3-TTA improves average Dice by 10.40 and 17.68 percentage points over the source-only baseline on heart and prostate segmentation, respectively, and consistently outperforms state-of-the-art TTA methods across six target domains.} {For natural images, we further evaluated our method on the Adverse Conditions Dataset for Cityscapes with DeepLabV3+, obtaining an improvement of Dcie by 16.90 percentage points over the source model.} 
{In continual TTA settings, where the model is adapted sequentially across multiple target domains, A3-TTA maintains high performance with strong resistance to forgetting.}

\begin{figure*}[t]
    \centering
\centerline{\includegraphics[width=18cm]{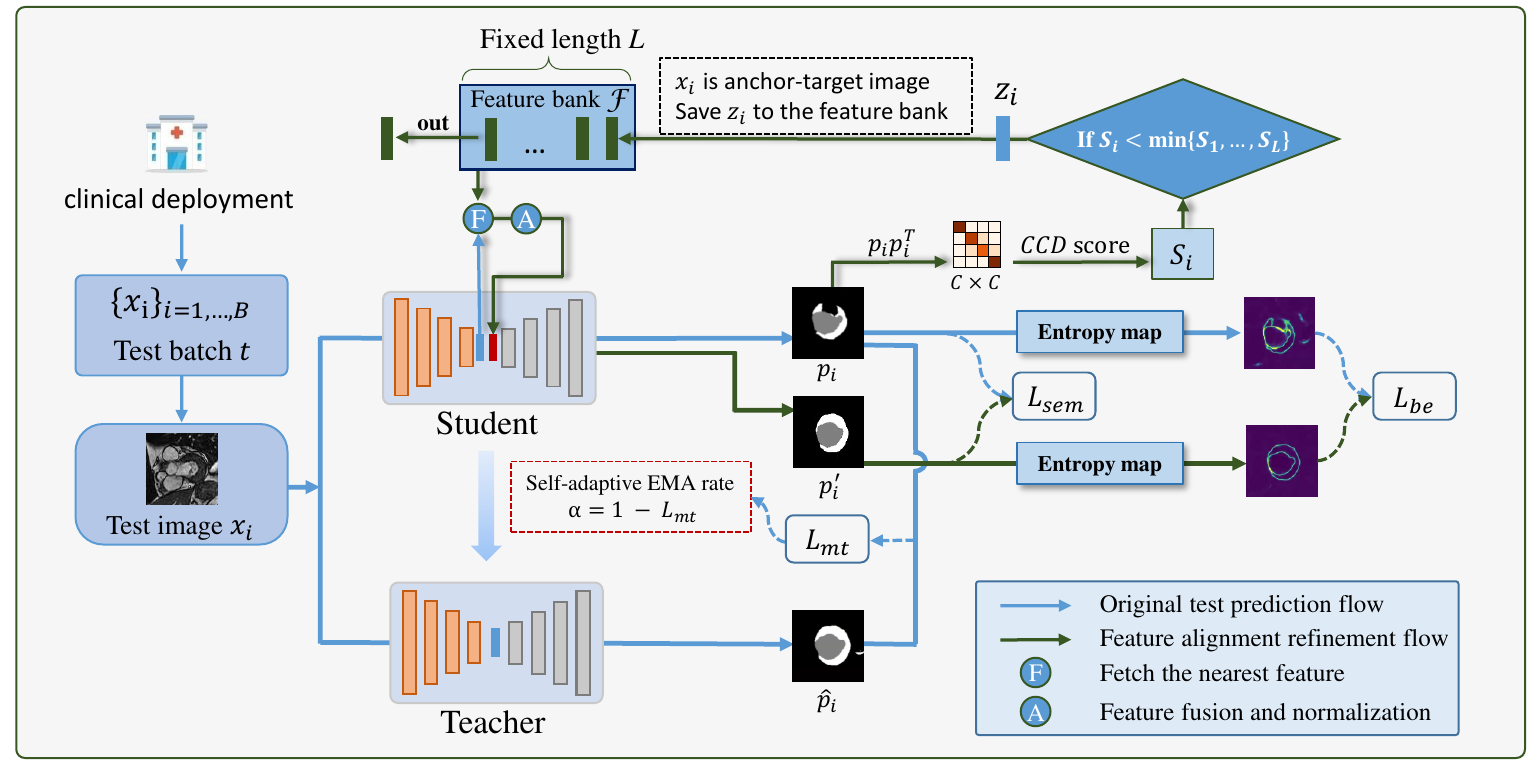}}
    \caption{Overview of our A3-TTA. A dynamic feature bank is constructed using anchor-target samples selected based on CCD. For a test image $x_i$, the most similar feature from the bank is fetched to obtain refined pseudo labels that are used for supervision via the semantic loss $L_{sem}$ and the boundary-guided entropy minimization loss $L_{be}$. A self-adaptive mean teacher framework is also employed for more robust adaptation through the loss $L_{mt}$. After each iteration update, the student model generates the final predictions for the current test images. }
    \label{fig:method}
\end{figure*}

\section{Related Works}
\subsection{Unsupervised Domain Adaptation}
{UDA} transfers knowledge from labeled source-domain data to an unlabeled target domain~\cite{ye2022online,kumari2024deep,zheng2024dual,cai2025style,peng2024unsupervised,hou2016unsupervised}. Current UDA approaches focus on three main strategies. First, image appearance alignment transforms target-domain images to resemble source-domain images~\cite{tzeng2017adversarial,dorent2023crossmoda,xu2023novel}, reducing the domain gap through techniques like image translation. For example, OMUDA~\cite{xu2023novel} disentangles content and style to translate source images into multiple target domains while maintaining cross-modality structural consistency through style constraints. Second, feature alignment minimizes the difference between source and target feature distributions, promoting domain-invariant representations~\cite{chen2020unsupervised}. Wu et al.~\cite{wu2021unsupervised} employ dual Variational Auto-Encoders (VAEs) to align features across domains by driving them toward a shared Gaussian distribution, {while Deng et al.~\cite{deng2022dynamic} proposed a dynamic instance-level alignment strategy that improves feature alignment granularity across domains}. Thirdly, output alignment generates pseudo-labels for target domain data, which are refined through self-training. FPL+~\cite{wu2024fpl-plus} generates and filters high-quality pseudo-labels for target domain images, enabling better training of the target segmentation model.
Despite the effectiveness of these methods, they rely on access to source-domain images, which is often impractical due to privacy concerns, computational costs, and bandwidth limitations.

\subsection{Test-Time Adaptation}
TTA aims to adapt the source model using only unlabeled target data at test time, without requiring structural changes during pre-training~\cite{chen2025gradient,zhang2024pass,liu2024mlfa}. 
Additionally, TTA requires that the model should be updated on the fly, without reliance on multi-epoch fine-tuning~\cite{chen2024each}. Early works on TTA focused on classification tasks. For example, Sun et al.~\cite{sun2020test} introduced an auxiliary branch to predict rotation angles in the target domain, while Karani et al.\cite{karani2021test} proposed a shallow image normalization network fine-tuned using Denoising Auto-Encoder (DAE) outputs. However, these methods require prior modifications to the source model, limiting their applicability to general models not pre-designed for adaptation.
Without auxiliary tasks for supervision, 
PTBN~\cite{nado2020evaluating} utilizes current batch statistics in the batch normalization layer, replacing source statistics. TENT~\cite{wang2021tent} further fine-tunes batch normalization parameters through entropy minimization. {Yang et al.~\cite{yang2022dltta} introduced DLTTA to dynamically adjusts the learning rate during test-time training to enhance adaptation stability.} 
SAR~\cite{niu2023towards} stabilizes TTA by removing noisy samples with large gradients and promoting flat minima in model weights. 
To improve pseudo-label quality in segmentation tasks, 
UPL-TTA~\cite{wu2023upltta}  uses data augmentation and decoder ensembles, while CoTTA~\cite{wang2022continual} and Zhu et al.~\cite{zhu2024improving} employ mean-teacher frameworks for pseudo-label generation. {More recently, Tan et al.~\cite{tan2025uncertainty} proposed an uncertainty-calibrated TTA framework that explicitly models prediction confidence to reduce error accumulation and prevent forgetting.} However, when feature distributions are misaligned or the source distribution is unknown, both data augmentation and mean-teacher methods struggle with distribution alignment, resulting in unreliable pseudo-labels. 
{Outside image segmentation, UCCH~\cite{hu2022unsupervised} learns compact, discriminative embeddings by combining learnable hashing with momentum and a cross-modal ranking loss to reduce false negatives, and RCL~\cite{hu2023cross} tackles partially mismatched pairs via complementary contrastive learning and risk upper bounds. Differently from these works, we propose CCD as a new efficient metric for uncertainty estimation in  TTA for segmentation tasks, and propose similarity-based feature alignment and boundary-guided dense supervision to filter out noise for more robust TTA. }

\section{Method}\label{method}
We propose A3-TTA, a unified test-time adaptation framework that constructs a stable pseudo-supervision pathway via anchor-guided refinement. As shown in Fig.~\ref{fig:method}, given several consecutive batches of test samples, we identify high-confidence ATIs based on a class compactness criterion and store their latent features in a dynamic feature bank. For each test image, the most similar anchor feature is retrieved and fused to refine its representation, enabling improved pseudo-label generation. These labels guide adaptation through semantic and boundary-aware objectives. To ensure stable single-pass optimization, we adopt a mean-teacher framework with a self-adaptive EMA rate based on prediction consistency.

\subsection{Problem Definition}
Let ${f = g \circ h}$ denote a segmentation model, with $h$ serving as the feature extractor and $g$ as the prediction head, respectively. Let $f_{\theta_0}$ denote a model trained on the source domain data $\mathcal{X}_S$. Upon deployment, the model encounters test domain data $\mathcal{X}_T$, where  $\mathcal{X}_T$ and $\mathcal{X}_S$ may have different intensity or appearance distributions. In TTA setting~\cite{wang2021tent}, the model must adapt in real-time to each incoming test batch ${\mathcal{B}_{t}} = \{ x_i\}_{i=1}^B$, with $t$ indicating the $t$-th batch, with a batch size of $B$. This process requires updating the model parameters from ${\theta_{t-1}}$ to ${\theta_{t}}$ to progressively enhance prediction accuracy for the testing images in the $t$-th batch. Importantly, each batch undergoes the model update and final inference before moving on to the next, guaranteeing that the entire dataset is tested within a single epoch in TTA with an efficient update of the model.

\subsection{Anchor-Target Images }
For each test image \( x_i \) in a batch, we extract its latent feature \( \boldsymbol{z}_i = h(x_i) \) and generate softmax predictions \( \boldsymbol{p}_i = g(\boldsymbol{z}_i) \). Motivated by the observation in Fig.~\ref{fig:fig1_anchor} that the quality of \( \boldsymbol{p}_i \) has a large variation with some of the target domain images close to the distribution boundary of training images, we define these images as ATIs that are treated as from an intermediate domain that serves as a bridge between the source domain and target domain to assist the adaptation process on the other target-domain images. 

Building on the previous observation~\cite{saito2021tune,zhao2023towards} that the predictions of a well-adapted model should exhibit compactness, and unlike existing works~\cite{saito2021tune,zhao2023towards}  that use pixel-level compactness to estimate prediction confidence, 
we propose a new compactness score \( S_i \) based on class-level prediction compactness, termed Class Compact Density (CCD), for each test image. 
Unlike entropy- or MC Dropout–based uncertainty estimators, which are either calculated at pixel-wise 
or require multiple forward passes, our CCD is a global, class-level compactness signal, and only needs a  single forward pass for efficient calculation.

Specifically, for each test image \( x_i \) with a shape of \(W \times H \), we first reshape the softmax predictions \( \boldsymbol{p}_i \) from \( C \times W \times H \) to \( C \times N \), where \( N = W \times H \) is the total number of pixels, and \( C \) is the class number. The corresponding compactness score \( S_i \) is then calculated as:
\begin{equation}
S_i = -\sum \left( \mathrm{softmax}\left(\boldsymbol{p}_i \boldsymbol{p}_i^{\mathrm{T}}\right) \right) \log\left( \mathrm{softmax}\left(\boldsymbol{p}_i \boldsymbol{p}_i^{\mathrm{T}}\right) \right),
\end{equation}
where \( \boldsymbol{p}_i \boldsymbol{p}_i^{\mathrm{T}} \in \mathbb{R}^{C \times C} \) is the class-wise similarity matrix computed via the outer product of softmax predictions. A column-wise softmax is applied to normalize similarity scores, and the summation is taken over all \( C \times C \) elements. A lower \( S_i \) value indicates that the similarity distribution is more concentrated along the diagonal, reflecting higher prediction compactness and lower inter-class ambiguity. 

For each new test scenario, we initialize an anchor-target feature bank $\mathcal{F}$ with a fixed capacity $L$. 
The parameter $L$ controls the maximum number of anchor-target features that can be stored in the bank, thereby determining the diversity of anchor references available for alignment. 
A larger $L$ enables the bank to retain more diverse anchor-target samples, which can improve adaptation stability, but it also requires more memory and increases retrieval cost. 
During adaptation, the compactness score \( S_i \) of each test image \( \boldsymbol{x}_i \) is compared against the highest score in \( \mathcal{F} \). In early iterations, when \( \mathcal{F} \) is not yet full, the top half of the current batch is directly added. 
Once $\mathcal{F}$ is full, we keep the $L$ smallest scores: letting $S_{\max}$ be the largest score  in the current $\mathcal{F}$, if a new sample has $S_i < S_{\max}$, we take it as an ATI and replace the entry attaining $S_{\max}$ with $\boldsymbol{z}_i$. This ensures \( \mathcal{F} \) remains aligned with the evolving distribution. 

\subsection{Feature Alignment-based Pseudo Label Refinement}
To address the instability and low quality of pseudo labels caused by heuristic perturbation-based strategies in existing TTA approaches, we introduce a Feature Alignment-based Refinement (FAR) strategy to improve the quality of dense pseudo labels. {For each test image \( x_i \), we calculate the cosine similarity between its latent feature \( \boldsymbol{z}_i \) and each feature in the anchor-target feature bank $\mathcal{F}$}, selecting the most similar anchor-target feature \( \boldsymbol{z}'_i \) from the bank:
\begin{equation}
    \boldsymbol{z}'_i = \underset{\boldsymbol{z} \in \mathcal{F}}{\arg\max} \frac{\boldsymbol{z} \cdot \boldsymbol{z}_i}{\|\boldsymbol{z}\| \|\boldsymbol{z}_i\|},
\end{equation}
{where $\boldsymbol{z}_i$ and each $\boldsymbol{z}\!\in\!\mathcal{F}$ are {flattened} bottleneck feature vectors in $\mathbb{R}^{D}$. }
The anchor-target feature \( \boldsymbol{z}'_i \) plays a crucial role in minimizing domain disparity by aligning target features with source-like representations, serving as a reliable reference to guide the adaptation of less certain or misaligned test images. To leverage the robust characteristics of these anchor-target features to improve feature alignment and reduce domain shift, our FAR strategy contains a similarity-aware feature fusion step and an anchor-based feature normalization step:

\textbf{1. Similarity-aware Feature Fusion:} The first step is to fuse \( \boldsymbol{z}_i \) with \( \boldsymbol{z}'_i \) to obtain a fused feature $\boldsymbol{z}^*_i$:
\begin{equation}
    \boldsymbol{z}^*_i = (1 - \lambda) \cdot \boldsymbol{z}_i + \lambda \cdot \boldsymbol{z}'_i,
\end{equation}
\begin{equation}
    \lambda \;=\; \max\!\left(0,\; \frac{\boldsymbol{z}_i \cdot \boldsymbol{z}'_i}{\|\boldsymbol{z}_i\| \,\|\boldsymbol{z}'_i\|}\right),
\end{equation}
{where $\lambda\in[0,1]$ is a scalar that serves as the coefficient for image-level fusion. The $max$ operation makes sure that only samples with positive correlations are considered for fusion.}
This fusion ensures that the resulting feature \( \boldsymbol{z}^*_i \) balances the properties of both inputs, enhancing the model's resilience to domain shift. {Notably, the fusion is applied only to the bottleneck layer in the encoder, while all other encoder layers and skip connections remain untouched. This design allows semantic alignment at a global level while preserving spatial structure and fine details from the original input.}

\textbf{2. Anchor-based Feature Normalization:} After fusion, we then normalize \( \boldsymbol{z}^*_i \) using the mean and standard deviation of the anchor-target feature \( \boldsymbol{z}'_i \). This step ensures that the distribution of the fused feature aligns more closely with that of the anchor-target feature, helping to reduce domain discrepancies. The normalized feature \( \hat{\boldsymbol{z}}_i \) is calculated as:
\begin{equation}
    \hat{\boldsymbol{z}}_i = \frac{\boldsymbol{z}^*_i - \mu(\boldsymbol{z}'_i)}{\sigma(\boldsymbol{z}'_i)\,{+\;\varepsilon}},
\end{equation}
{where \(\mu(\cdot)\) and \(\sigma(\cdot)\) are computed over the flattened vector of $z'_i$, and \(\varepsilon=10^{-5}\) ensures numerical stability.}

Finally, the normalized feature \( \hat{\boldsymbol{z}}_i \) is fed into the model's decoder to generate a refined pseudo label \( \boldsymbol{p}'_i = g(\hat{\boldsymbol{z}}_i) \) for the test image \( x_i \). This refinement process enhances the dense supervision during test-time adaptation, resulting in more accurate and robust predictions in the target domain.

\subsection{Boundary-Guided Dense Supervision}
To maximize the effectiveness of refined pseudo labels \( \boldsymbol{p}'_i \) generated via anchor-guided alignment, we introduce a dual loss strategy that provides reliable dense supervision by incorporating both semantic consistency and boundary-guided entropy minimization. This approach ensures a comprehensive alignment between the original and refined predictions at the pixel level, enhancing the overall model performance.

First, to ensure semantic consistency across all pixels, we employ a normalized cross-entropy loss~\cite{bishop2006pattern}, which measures the alignment between the refined pseudo label \( \boldsymbol{p}'_i \) and the original prediction \( \boldsymbol{p}_i \):
\begin{equation}\label{eq_sem}
\mathcal{L}_{sem}(\boldsymbol{p}'_i, \boldsymbol{p}_i) = -\frac{1}{N\log_2(C)} \sum_{n=1}^{N}\sum_{c=1}^{C} \boldsymbol{p}'_{i,n,c} \log(\boldsymbol{p}_{i,n,c}),
\end{equation}
where \( N \) is the total number of pixels, \( \boldsymbol{p}_{i,n,c} \) denotes the predicted probability for class \( c \) at pixel \( n \) in image \( x_i \). After normalization by \( \log_2(C) \), the value range of \( \mathcal{L}_{sem} \) is \([0, 1]\).

Second, to effectively reduce prediction uncertainty while avoiding the model degradation often caused by unsupervised entropy minimization~\cite{vu2019advent}, we introduce a boundary-guided entropy minimization loss. This loss operates on every pixel, focusing on areas with high entropy typically concentrated near object boundaries~\cite{tu2017skeletal}. By leveraging these boundary cues to guide entropy minimization, we provide dense supervision that reduces uncertainty specifically at object boundaries, enhancing boundary clarity and overall segmentation performance:
\begin{equation}
\mathcal{L}_{be}(\boldsymbol{p}'_i, \boldsymbol{p}_i) = \frac{1}{N}\sum_{n=1}^{N} \left| E(\boldsymbol{p}_{i,n}) - E(\boldsymbol{p}'_{i,n}) \right|,
\end{equation}
where \( E(\boldsymbol{p}_{i,n}) \) represents the entropy value at pixel \( n \), indicating uncertainty in the prediction, calculated as:
\begin{equation}\label{eq_entropy}
    E(\boldsymbol{p}_{i,n}) = -\sum_{c=1}^{C} \boldsymbol{p}_{i,n,c} \log(\boldsymbol{p}_{i,n,c}),
\end{equation}
where \( \boldsymbol{p}_{i,n,c} \) is the predicted probability for class \( c \) at pixel \( n \) in test image \( x_i \).

\begin{table*}[t]
  \centering
  \caption{ Comparison between different TTA methods on the M\&MS Dataset in terms of Dice (\%). \dag denotes a significant improvement (p-value $<$ 0.05) over the best existing method.}
  \begin{adjustbox}{width=1.0\textwidth}
    \begin{tabular}{l|ccc|ccc|ccc|c}
    \hline
    \multirow{2}[4]{*}{Method} & \multicolumn{3}{c|}{Domain B} & \multicolumn{3}{c|}{Domain C} & \multicolumn{3}{c|}{Domain D} & \multirow{2}[4]{*}{Average} \bigstrut\\
\cline{2-10}          & LV    & MYO   & RV    & LV    & MYO   & RV    & LV    & MYO   & RV    &  \bigstrut\\
    \hline
    Source Only & 80.89±23.46 & 66.81±20.29 & 67.47±35.35 & 75.56±27.18 & 56.99±23.46 & 60.54±37.16 & 83.84±22.72 & 69.36±21.35 & 64.51±37.40 & 69.72±27.38 \bigstrut[t]\\
    PTBN~\cite{nado2020evaluating}  & 83.93±20.82 & 74.72±15.62 & 68.38±34.75 & 82.21±21.11 & 71.46±17.18 & 65.28±35.57 & 84.56±20.74 & 72.39±19.24 & 66.34±35.99 & 74.62±24.31 \\
    TENT~\cite{wang2021tent}  & 83.92±22.64 & 74.11±18.39 & 71.32±33.63 & 82.76±21.57 & 71.76±18.37 & 67.46±35.20 & 84.47±21.62 & 72.17±20.28 & 68.17±35.66 & 75.41±25.13 \\
    MT~\cite{tarvainen2017mean}    & 83.54±22.18 & 74.64±16.97 & 70.01±34.52 & 81.72±22.29 & 71.35±17.83 & 66.54±35.64 & 83.91±22.24 & 72.24±19.94 & 67.15±36.51 & 74.88±25.10 \\
    CoTTA~\cite{wang2022continual} & 84.88±19.19 & 76.22±13.87 & 70.52±34.33 & 82.13±21.65 & 72.16±17.22 & 66.59±35.60 & 84.88±20.34 & 72.91±18.63 & 68.17±35.12 & 75.77±23.62 \\
    SAR~\cite{niu2023towards} & 84.11±21.08 & 75.02±15.56 & 68.73±34.71 & 82.62±20.90&71.85±16.86&64.44±36.02&84.80±20.83&72.64±19.03&66.62±35.86&74.81±24.32 \\
    InTEnt~\cite{dong2024medical}   &81.34±23.02&67.77±19.78&67.84±35.09&76.55±26.50&58.71±22.78&61.05±36.89&84.05±22.41&69.81±21.16&65.14±37.11& 70.42±26.96    \\
    VPTTA~\cite{chen2024each}   &82.97±21.53&73.19±16.36&65.41±35.82&82.15±21.12&71.29±16.67&62.42±36.38&84.17±21.70&71.98±19.48&63.09±36.95& 73.16±24.91    \\
    {GraTa}~\cite{chen2025gradient}   &{82.88±21.64}&{71.77±16.92}&{63.76±35.53}&{81.17±22.02}&{68.24±18.72}&{60.85±36.29}&{82.88±22.67}&{69.81±20.30}&{60.89±36.78}& {71.67±20.74}    \\
    {EDCP}~\cite{liu2025efficient}   
& {80.55±24.00} & {65.83±20.89} & {67.02±35.42} 
& {77.57±25.07} & {58.98±22.75} & {62.59±36.63} 
& {83.30±23.31} & {68.62±21.56} & {64.61±37.31} & {69.94±21.09} \\
    A3-TTA & \textbf{87.58±17.64}$^{\dag}$ & \textbf{80.04±13.99}$^{\dag}$ & \textbf{78.80±29.79}$^{\dag}$ & \textbf{85.41±19.46}$^{\dag}$ & \textbf{75.77±17.60}$^{\dag}$ & \textbf{72.09±34.19}$^{\dag}$ & \textbf{86.77±19.37}$^{\dag}$ & \textbf{75.31±19.26}$^{\dag}$ & \textbf{73.70±33.59}$^{\dag}$ & \textbf{80.12±22.17}$^{\dag}$ \bigstrut[b]\\
    \hline
    \end{tabular}
    \end{adjustbox}
  \label{tab:mms_sota_dice}
\end{table*}

\begin{table*}[t]
  \centering
  \caption{ Comparison of different TTA methods on the M\&MS Dataset in terms of ASSD (mm). \dag denotes a significant improvement (p-value $<$ 0.05) over the best existing method.}
  \begin{adjustbox}{width=1.0\textwidth}
    \begin{tabular}{l|ccc|ccc|ccc|c}
    \hline
    \multirow{2}[4]{*}{Method} & \multicolumn{3}{c|}{Domain B} & \multicolumn{3}{c|}{Domain C} & \multicolumn{3}{c|}{Domain D} & \multirow{2}[4]{*}{Average} \bigstrut\\
\cline{2-10}          & LV    & MYO   & RV    & LV    & MYO   & RV    & LV    & MYO   & RV    &  \bigstrut\\
    \hline
    Source Only & 4.17±3.15 & 4.18±2.94 & 4.58±3.59 & 4.80±3.27 & 5.24±3.21 & 5.43±3.75 & 3.30±2.72 & 3.47±2.56 & 4.85±3.70 & 4.47±3.23 \bigstrut[t]\\
    PTBN~\cite{nado2020evaluating}  & 3.80±3.21 & 4.04±2.87 & 4.77±3.58 & 4.35±3.33 & 4.76±3.11 & 5.21±3.58 & 3.53±2.85 & 3.91±2.78 & 4.94±3.60 & 4.35±3.22 \\
    TENT~\cite{wang2021tent}  & 3.13±2.75 & 3.31±2.58 & 4.18±3.45 & 3.82±3.08 & 3.93±2.91 & 4.66±3.52 & 3.28±2.63 & 3.45±2.57 & 4.52±3.58 & 3.75±2.99 \\
    MT~\cite{tarvainen2017mean}    & 3.57±3.11 & 3.76±2.79 & 4.37±3.56 & 4.16±3.29 & 4.56±3.11 & 4.81±3.56 & 3.44±2.85 & 3.71±2.74 & 4.64±3.62 & 4.08±3.18 \\
    CoTTA~\cite{wang2022continual} & 3.50±3.01 & 3.81±2.76 & 4.39±3.51 & 4.21±3.32 & 4.48±3.12 & 4.97±3.61 & 3.40±2.76 & 3.82±2.75 & 4.65±3.54 & 4.10±3.15 \\
    SAR~\cite{niu2023towards} & 3.63±3.09&4.00±2.82&4.64±3.53&4.20±3.27&4.59±3.05&5.27±3.57&3.41±2.77&3.81±2.70&4.84±3.57&4.25±3.16 \\
    InTEnt~\cite{dong2024medical} & 4.10±3.13&4.10±2.90&4.54±3.57&4.69±3.26&5.11±3.18&5.36±3.71&3.27±2.70&3.43±2.54&4.77±3.68&4.39±3.20 \\
    VPTTA~\cite{chen2024each} & 3.86±3.14&3.75±2.75&4.90±3.56&4.24±3.26&4.23±2.99&5.40±3.54&3.29±2.69&3.47±2.58&5.05±3.57&4.25±3.14 \\
    {GraTa}~\cite{chen2025gradient}   &{4.34±3.45}&{5.62±3.15}&{5.80±3.59}&{4.94±3.52}&{6.26±3.12}&{6.37±3.55}&{4.23±3.22}&{5.28±3.05}&{6.27±3.54}& {5.43±2.68}    \\
    {EDCP}~\cite{liu2025efficient}   
& {4.14±3.14} & {4.22±2.96} & {4.62±3.60}
& {4.69±3.22} & {5.07±3.14} & {5.20±3.72} 
& {3.36±2.72} & {3.52±2.57} & {4.87±3.69} & {4.44±2.50} \\
    A3-TTA &  \textbf{2.49±2.16}$^{\dag}$ & \textbf{2.42±1.76}$^{\dag}$ & \textbf{3.62±3.21}$^{\dag}$ &  \textbf{2.87±2.42}$^{\dag}$ & \textbf{2.80±2.18}$^{\dag}$ & \textbf{3.81±3.35}$^{\dag}$ & \textbf{2.55±1.97}$^{\dag}$ & \textbf{2.59±2.02}$^{\dag}$ & \textbf{3.59±3.28}$^{\dag}$ & \textbf{2.95±2.46}$^{\dag}$ \bigstrut[b]\\
    \hline
    \end{tabular}
    \end{adjustbox}
  \label{tab:mms_sota_assd}
\end{table*}

\begin{table*}[t]
  \centering
  \caption{Comparison between different TTA methods on the prostate MRI Dataset. \dag denotes a significant improvement (p-value $<$ 0.05) over the best existing method.}
    \begin{tabular}{l|ccc|c|ccc|c}
    \hline
    \multirow{2}[4]{*}{Method} & \multicolumn{4}{c|}{Dice (\%)} & \multicolumn{4}{c}{ASSD (mm)} \bigstrut\\
\cline{2-9}          & Domain D     & Domain E     & Domain F     & Average & Domain D     & Domain E     & Domain F     & Average \bigstrut\\
    \hline
    Source Only & 83.04±16.19 & 37.33±30.21 & 72.33±25.01 & 63.39±31.64 & 2.45±1.87 & 15.33±16.17 & 5.10±8.86 & 7.87±12.28 \bigstrut[t]\\
    PTBN~\cite{nado2020evaluating}  & 82.60±17.06 & 58.40±27.70 & 76.87±20.90 & 72.18±24.83 & 4.92±6.07 & 17.87±12.79 & 8.03±9.25 & 10.51±11.35 \\
    TENT~\cite{wang2021tent}  & 83.57±16.19 & 66.69±25.47 & 80.25±16.62 & 76.52±21.41 & 3.20±2.83 & 6.60±3.46 & 3.88±3.07 & 4.62±3.48 \\
    MT~\cite{tarvainen2017mean}    & 82.11±17.74 & 59.28±28.11 & 75.94±22.25 & 72.03±25.23 & 3.68±3.22 & 7.64±3.21 & 4.80±3.40 & 5.45±3.68 \\
    CoTTA~\cite{wang2022continual} & 83.22±16.56 & 62.95±26.83 & 78.08±19.21 & 74.38±23.21 & 3.42±2.99 & 7.28±3.37 & 4.67±3.32 & 5.19±3.62 \\
    SAR~\cite{niu2023towards} & 84.22±15.05 & 62.51±25.71 & 78.47±20.38 &   74.67±22.98  & 2.75±2.18&8.19±2.38&3.90±2.61 & 5.05±3.37 \\
    InTEnt~\cite{dong2024medical} &83.24±15.91&39.88±30.20&73.34±24.39&64.67±30.83   & 2.41±1.72 &7.39±2.81 &3.48±2.29& 4.52±3.18   \\
    VPTTA~\cite{chen2024each} &83.65±15.98&42.99±31.09&73.84±23.38&66.07±23.71   & 2.87±2.47 &7.45±2.93 &3.88±2.64& 4.81±2.68   \\
    {GraTa}~\cite{chen2025gradient} 
& {79.02±19.35} & {51.05±27.78} & {71.11±23.11} & {66.56±26.63}   
& {5.61±3.70} & {9.25±1.84} & {7.69±3.10} & {7.57±3.31}   \\
{EDCP}~\cite{liu2025efficient} 
& {83.05±16.19} & {37.35±30.22} & {72.34±25.02} & {63.40±31.64}   
& {2.44±1.74} & {7.65±2.78} & {3.62±2.40} & {4.67±3.27}   \\
    A3-TTA &  
    \textbf{84.78±15.16}$^{\dag}$ & \textbf{75.56±19.91}$^{\dag}$ & \textbf{83.43±14.05}$^{\dag}$ & \textbf{81.07±17.19}$^{\dag}$ & \textbf{2.40±2.00} & \textbf{4.35±2.99}$^{\dag}$ & \textbf{2.47±1.60}$^{\dag}$ & \textbf{3.11±2.48}$^{\dag}$ \bigstrut[b]\\
    \hline
    \end{tabular}
  \label{tab:prostate_assd_dice}%
\end{table*}%

\subsection{Self-Adaptive Robust Mean Teacher}
Although the FAR strategy generates reliable pseudo labels, achieving robust and consistent performance in a single-epoch adaptation scenario with dynamically arriving test samples remains challenging. Adapting the model on-the-fly without stable and consistent guidance can lead to optimization instability, particularly when the distribution of test data undergoes abrupt or significant shifts.

To address these issues, we employ a Mean Teacher framework to enhance the stability of TTA. Let \( \theta_t \) and \( \hat{\theta}_t \) denote the parameters of the student and teacher models at iteration \( t \), respectively. Traditional methods~\cite{tarvainen2017mean,chen2020multi} update the teacher model using a constant Exponential Moving Average (EMA) rate: \( \hat{\theta}_{t} = \alpha \hat{\theta}_{t-1} + (1 - \alpha) \theta_{t} \), with a fixed \( \alpha \), typically 0.99.

However, in single-epoch adaptation with dynamic test batches, a fixed \( \alpha \) cannot handle rapid changes in data distribution. Therefore, we introduce a Self-Adaptive EMA Rate (SER) that dynamically adjusts the teacher update rate based on observed distribution shifts.

We quantify the distribution shift by calculating the divergence between the teacher's prediction \( \hat{\boldsymbol{p}}_i = f_{\hat{\theta}_{t-1}}(x_i) \) and the student's prediction \( \boldsymbol{p}_i = f_{\theta_t}(x_i) \), using a normalized cross-entropy loss:
\begin{equation}\label{eq_mt}
\mathcal{L}_{mt}(\hat{\boldsymbol{p}}_i, \boldsymbol{p}_i) = -\frac{1}{N\log_2(C)} \sum_{n=1}^{N}\sum_{c=1}^{C} \hat{\boldsymbol{p}}_{i,n,c} \log(\boldsymbol{p}_{i,n,c}),
\end{equation}
where \( \hat{\boldsymbol{p}}_{i,n,c} \) and \( \boldsymbol{p}_{i,n,c} \) are the predicted probabilities for class \( c \) at pixel \( n \). The normalized loss \( \mathcal{L}_{mt} \in [0, 1] \) indicates the discrepancy between the teacher and student models.

The teacher's parameters are updated using:

\begin{equation}
\hat{\theta}_{t} = (1 - \mathcal{L}_{mt}) \hat{\theta}_{t-1} + \mathcal{L}_{mt} \theta_{t},
\end{equation}
where when the divergence \( \mathcal{L}_{mt} \) is large, the teacher model aligns more with the student model by placing greater weight on \( \theta_{t} \). Conversely, when \( \mathcal{L}_{mt} \) is small, the teacher retains more of its previous parameters \( \hat{\theta}_{t-1} \), maintaining stability.
By dynamically adjusting the EMA rate, our SER framework balances rapid adaptation to new distributions with model stability, which is crucial in single-epoch adaptation.

The overall loss function in our A3-TTA is:
\begin{equation}
\mathcal{L}_{TTA} = \mathcal{L}_{sem}(\boldsymbol{p}'_i, \boldsymbol{p}_i) + \beta\, \mathcal{L}_{be}(\boldsymbol{p}'_i, \boldsymbol{p}_i) + \gamma\, \mathcal{L}_{mt}(\hat{\boldsymbol{p}}_i, \boldsymbol{p}_i),
\end{equation}
where \( \beta \) and \( \gamma \) balance the contributions of the semantic, boundary, and mean teacher losses.

After computing \( \mathcal{L}_{TTA} \), we update the student model's parameters \( \theta_t \) via backpropagation. The student model then generates final predictions for the current batch of test images.

\begin{figure*}[t]
	\centering
	\centerline{\includegraphics[width=18cm]{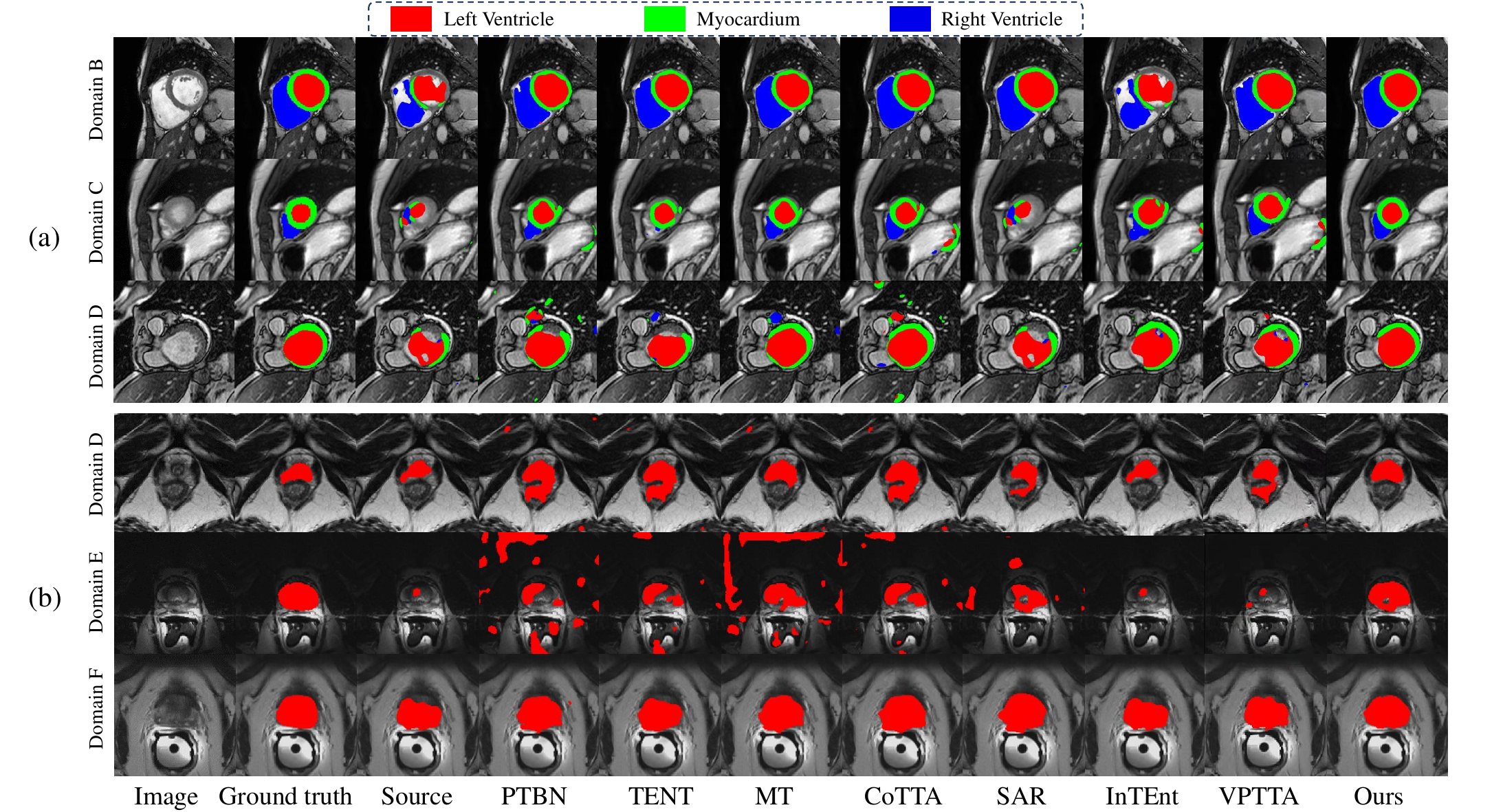}}
   \caption{Qualitative comparison of different TTA methods for (a) cardiac image segmentation and (b) prostate MRI segmentation across multiple target domains. 
  } \label{fig:segmentation}
\end{figure*}
\section{Experiment and Results}
\subsection{Datasets and Implementation Details}
{We used three multi-domain datasets for experiments, where two of them are from medical domains, and one is from natural image domains.}
\subsubsection{Cardiac Image Segmentation Dataset (M\&MS)}
It consists of 345 3D cardiac MRI volumes collected by 
imaging devices from four different scanner vendors~\cite{campello2021multi}: Siemens, Philips, General Electric, and Canon. The four scanner vendees were treated as four different domains. 
The slice number per volume varied between 10 and 13, with an in-plane resolution ranging from 0.85 to 1.45 mm and a slice thickness of 9.2 to 10 mm. Following the settings in~\cite{tomar2022opttta}, Domain A was used as the source domain, while Domains B, C, and D were set as the target domains respectively. The target tissues for segmentation include the Left Ventricle (LV), Right Ventricle (RV), and Myocardium (MYO).

\subsubsection{Prostate MRI Segmentation Dataset}
It consists of 3D MRI volumes sourced from six different sites (A to F), collected from the NIC-ISBI13~\cite{bloch2015nci}, I2CVB~\cite{lemaitre2015computer}, and PROMISE12~\cite{litjens2014evaluation} datasets. The number of cases for each site is 30 for sites A and B, 19 for site C, and 13, 12, and 12 for sites D, E, and F, respectively. These MRI scans were acquired using different imaging protocols, leading to heterogeneous data distributions. Following~\cite{tomar2022opttta}, data from site C were excluded due to poor quality. The union of sites A and B was used as the source domain, while sites D, E, and F served as the target domains. The slice thickness across these domains ranges from 1.25 to 4 mm. In line with the TTA setting~\cite{chen2024each}, all images in each target domain were utilized for adaptation and testing.

\subsubsection{{Adverse Conditions Dataset for Cityscapes}}
{It includes 4,006 urban images under four weather conditions (fog, nighttime, rain, and snow) with dense annotations of 19 classes~\cite{sakaridis2021acdc}. The nighttime condition has 1006 images, and each of the other conditions has 1000 images. 
We used the union of fog and nighttime as the source domain, and the other two conditions as target domains for TTA. } 

\subsubsection{Implementation Details}
To ensure fair comparison, all methods, including our approach and the baselines, are implemented using the same backbone, preprocessing pipeline, and training setup. For the two medical image datasets, we adopt the 2D U-Net~\cite{ronneberger2015u} as the segmentation backbone rather than a 3D network, owing to the large inter-slice spacing. Each slice was resized to $320\times320$ pixels and linearly normalized to $[-1,1]$ via min–max scaling. {For the Adverse Conditions Dataset for Cityscapes, we use DeepLabV3+ with a ResNet-101 encoder~\cite{chen2018encoder} as the segmentation model, and images were resized to $512\times256$}. 
The model was trained in the source domain for 200 epochs using the Dice loss and Adam optimizer with learning rate of $1.0 \times 10^{-3}$, and the checkpoint yielding the best performance on the source domain validation set was used to initialize test-time adaptation.
During adaptation, we set the batch size to 10 2D images, employing a single epoch of TTA with Adam optimizer with fixed learning rate of $1.0 \times 10^{-4}$ on all learnable parameters. Hyper-parameter optimization, conducted through experiments on domain B of the M\&MS dataset, resulted in setting $L = 40$, $\beta = 5$, $\gamma = 1$. The experiments were conducted with
 PyTorch 1.8.1 and NVIDIA GeForce RTX 2080Ti GPU. Evaluation metrics include the Dice Similarity Coefficient (DSC) and Average Symmetric Surface Distance (ASSD) for {medical image segmentation}, {and DSC and mean Intersection over Union (mIoU) for the Adverse Conditions Dataset for Cityscapes.}

\begin{table}[t]
  \centering
  \caption{{Comparison between different TTA methods on the Adverse Conditions Dataset for Cityscapes using DeepLabV3+.
  \dag denotes a significant improvement ($p<0.05$) over the best existing method.}}
    \begin{adjustbox}{width=0.5\textwidth}
    \begin{tabular}{l|cc|cc}
    \hline
    \multirow{2}[4]{*}{Method} & \multicolumn{2}{c|}{Dice (\%)} & \multicolumn{2}{c}{mIoU (\%)} \bigstrut\\
\cline{2-5}          & Rain     & Snow  & Rain & Snow  \bigstrut\\
    \hline
    Source Only & 42.01±10.77 & 39.40±8.79 & 37.53±10.63 & 34.87±8.75  \bigstrut[t]\\
    TENT~\cite{wang2021tent}  & 54.59±10.95 & 55.57±9.43 & 50.72±11.05 & 52.59±9.41  \\
    CoTTA~\cite{wang2022continual} & 45.83±10.93 & 44.34±10.18 & 41.31±10.80 & 39.71±10.12 \\
    SAR~\cite{niu2023towards} & 46.22±11.26 & 44.54±10.05 & 41.70±11.12 & 39.98±9.96 \\
    A3-TTA &  
    \textbf{56.73±10.17}$^{\dag}$ & \textbf{58.50±9.13}$^{\dag}$ & \textbf{53.00±10.34}$^{\dag}$ & \textbf{55.16±9.11}$^{\dag}$ \bigstrut[b]\\
    \hline
    \end{tabular}
    \end{adjustbox}
  \label{tab:acdccity_assd_dice}%
\end{table}%
\begin{table*}[htbp]
  \centering
  \caption{Ablation study of A3-TTA on Domain B of the M\&MS dataset. CCD=$\diamond$ means using entropy for filtering anchor-target images, SER=$\diamond$ means setting $\alpha$ to a fixed value of 0.99 in mean teacher.}
    \begin{tabular}{ccccc|ccc|c}
    \hline
    $\mathcal{L}_{mt}$  & $\mathcal{L}_{sem}$ & $\mathcal{L}_{be}$ & CCD   & SER   & LV    & MYO   & RV    & Average \bigstrut\\
    \hline
          &       &       &       &       & 80.89±23.46 & 66.81±20.29 & 67.47±35.35 & 71.72±20.24 \bigstrut[t]\\
    \checkmark    &       &       &       &    $\diamond$   & 83.49±22.12 & 74.52±17.07 & 69.84±34.60 & 75.95±19.27 \\
    \checkmark    & \checkmark    &       &    $\diamond$   &    $\diamond$   & 86.55±18.25 & 77.64±14.11 & 72.77±33.09 & 78.98±16.41 \\
    \checkmark    & \checkmark    & \checkmark    &   $\diamond$    &    $\diamond$   & 86.84±17.87 & 77.95±14.16 & 73.35±32.46 & 79.38±16.27 \\
    \checkmark    & \checkmark    & \checkmark    & \checkmark    &    $\diamond$   & 86.95±18.63 & 78.44±16.00 & 75.68±31.34 & 80.36±15.97 \\
    \checkmark    & \checkmark    & \checkmark    & \checkmark    & \checkmark    & \textbf{87.58±17.64} & \textbf{80.04±13.99} & \textbf{78.80±29.79} & \textbf{82.14±14.92} \bigstrut[b]\\
    \hline
    \end{tabular}%
  \label{tab:ablation}%
\end{table*}%

\subsection{Comparison with State-of-the-art Methods}
Our A3-TTA was compared with {nine} state-of-the-art TTA methods: 
1) \textbf{PTBN}~\cite{nado2020evaluating}, 
2) \textbf{TENT}~\cite{wang2021tent}, 
3) \textbf{MT}~\cite{tarvainen2017mean} that utilizes a teacher-student structure with an EMA rate of 0.99 for adaptation and updates all parameters, 
4) \textbf{CoTTA}~\cite{wang2022continual} that updates all parameters within a mean-teacher framework, employing test-time augmentation-based pseudo labels for adaptation;
5) \textbf{SAR}~\cite{niu2023towards} that uses sharpness-aware entropy minimization for adaptation; 
6) \textbf{InTEnt}~\cite{dong2024medical} that integrates predictions using weighted entropy statistics across various target domain estimates;
7) \textbf{VPTTA}~\cite{chen2024each} that uses low-frequency visual prompts to adapt each test image across continual target domains to a pre-trained model without modifying its parameters;
{8) \textbf{GRATA}~\cite{chen2025gradient} that regularizes test-time adaptation with gradient agreement, leveraging multiple augmented views of the same target image to stabilize optimization;}
{9) \textbf{EDCP}~\cite{liu2025efficient} that employs deformable convolutional prompts with an offset bank for parameter-efficient continual adaptation.} 
Our method was also compared to `\textbf{Source Only}', which directly applies the source model for inference on target domain datasets.

\subsubsection{Result for Cardiac Image Segmentation}
Table~\ref{tab:mms_sota_dice} and Table~\ref{tab:mms_sota_assd} present the quantitative comparison of the methods in terms of Dice and ASSD results for cardiac image segmentation. The source model achieved an average Dice score of 69.72\% across the three target domains. All existing TTA methods demonstrated improvements over the source model, with average Dice scores ranging from 70.42\% to 75.77\%. For instance, in Target Domain B, TENT~\cite{wang2021tent} increased the average Dice score for the LV from 80.89\% to 83.93\%, with an overall average Dice score of 74.62\% across all three domains. Other TTA methods, such as VPTTA~\cite{chen2024each},  CoTTA~\cite{wang2022continual} and SAR~\cite{niu2023towards}, further increased the average Dice scores to 73.16\%, 75.77\% and 74.81\%, respectively. In contrast, our A3-TTA method showed a substantial improvement, with an average Dice score increase of 10.40 percentage points over the source model, outperforming the best existing methods.
Additionally, all compared TTA methods showed improvements over the source model in terms of ASSD. Our approach achieved an average ASSD of 2.95 mm, significantly outperforming the other methods. The qualitative comparison in Fig.~\ref{fig:segmentation} (a) illustrates that while existing methods often result in under-segmentation in the cardiac image segmentation dataset, our A3-TTA method excels in accurately delineating the target regions across different target domains.

\begin{figure*}[t]
	\centering
	\centerline{\includegraphics[width=18cm]{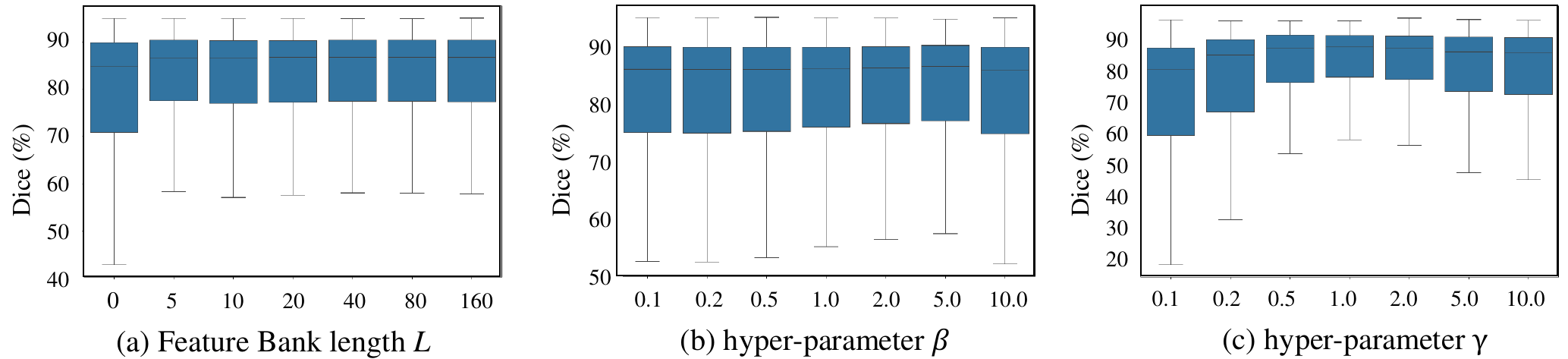}}
   \caption{Sensitivity analysis of hyper-parameters on Domain B of the M$\&$M dataset.
  } \label{fig:hypa}
\end{figure*}

\subsubsection{Result for Prostate MRI Segmentation}
We further evaluated the performance of the compared methods on the prostate MRI segmentation dataset. The quantitative results are presented in Table~\ref{tab:prostate_assd_dice}. The `Source Only' model only achieved an average Dice of 63.39\% across the target domains, while all the compared TTA methods showed improvements, with average Dice scores across the three target domains ranging from 64.67\% to 76.52\%. In contrast, our method significantly outperformed the others, achieving an average Dice score of 81.07\%.
In terms of ASSD, PTBN~\cite{nado2020evaluating} showed an increase compared to `Source Only', rising from 7.87 mm to 10.51 mm, while TENT~\cite{wang2021tent}, MT~\cite{tarvainen2017mean}, CoTTA~\cite{wang2022continual}, and SAR~\cite{niu2023towards} reduced it to 4.62 mm, 5.45 mm, 5.19 mm, and 5.05 mm, respectively, with InTEnt~\cite{dong2024medical} further lowering the ASSD to 4.52 mm. Notably, our approach achieved a significantly lower average ASSD of 3.11 mm, outperforming all the other methods.
The qualitative comparison in Fig.~\ref{fig:segmentation}(b) demonstrates that while existing methods often result in under-segmentation of the prostate in different target domains, our A3-TTA method excels in accurately delineating the target regions across different target domains.

\subsubsection{{Results on the Adverse Conditions Dataset for Cityscapes }}
{
We compared our A3-TTA against the top 3 existing TTA methods (TENT~\cite{wang2021tent}, CoTTA~\cite{wang2022continual}, SAR~\cite{niu2023towards}) according to Table~\ref{tab:mms_sota_dice} and~\ref{tab:prostate_assd_dice}.   
As shown in Table~\ref{tab:acdccity_assd_dice}, 
TENT~\cite{wang2021tent} outperformed the other two exiting methods, with a Dice of 54.59\% and 55.57\% for rain and snow domains, respectively. In contrast, our method improved the average Dice by 2.14 and 2.93 percentage points from TENT in the two target domains, respectively. 
The results in Table~\ref{tab:mms_sota_dice}, \ref{tab:prostate_assd_dice} and~\ref{tab:acdccity_assd_dice} show that our method generalizes to different types of segmentation tasks (both medical and natural images) under different segmentation backbones (U-Net and DeepLabV3+). 
}

\begin{figure*}[t]
    \centering
    \centerline{\includegraphics[width=18cm]{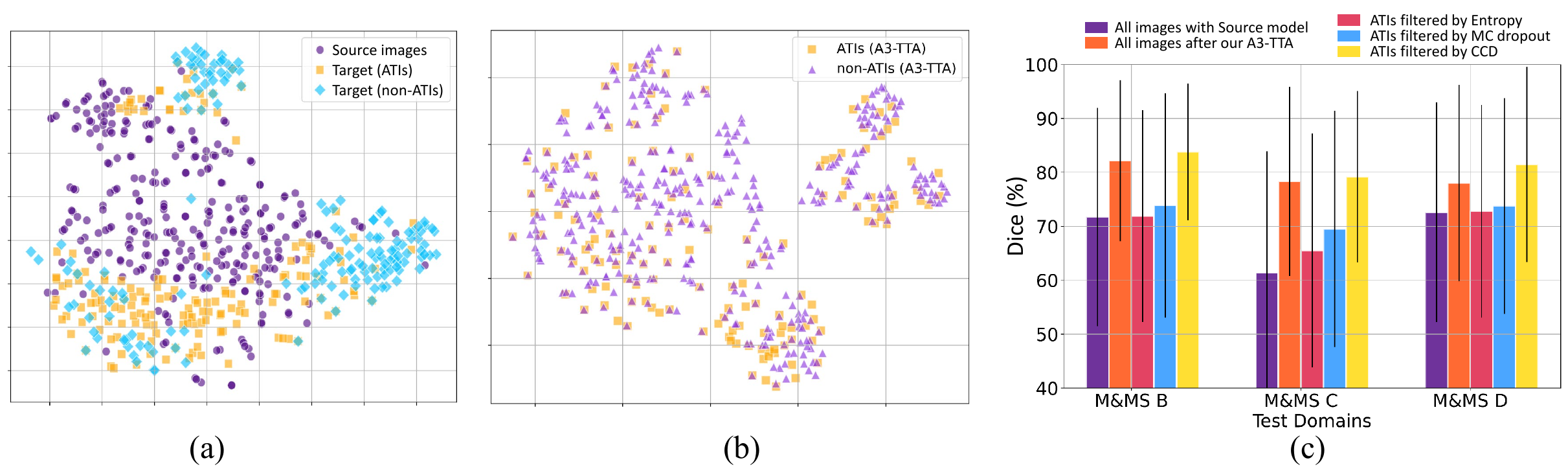}}
     \caption{Effectiveness of Anchor-Target Images (ATIs ). (a) T-SNE visualization of features of source images and target images (ATIs and non-ATIs) obtained by the source model. (b) T-SNE visualization of feature alignment between non-ATIs and ATIs after applying our A3-TTA. (c) Performance comparison between before and after A3-TTA, and quality of ATIs filtered by different methods.  }
    \label{fig:tsne_2}
\end{figure*}

\begin{figure}[t]
    \centering
    \centerline{\includegraphics[width=9cm]{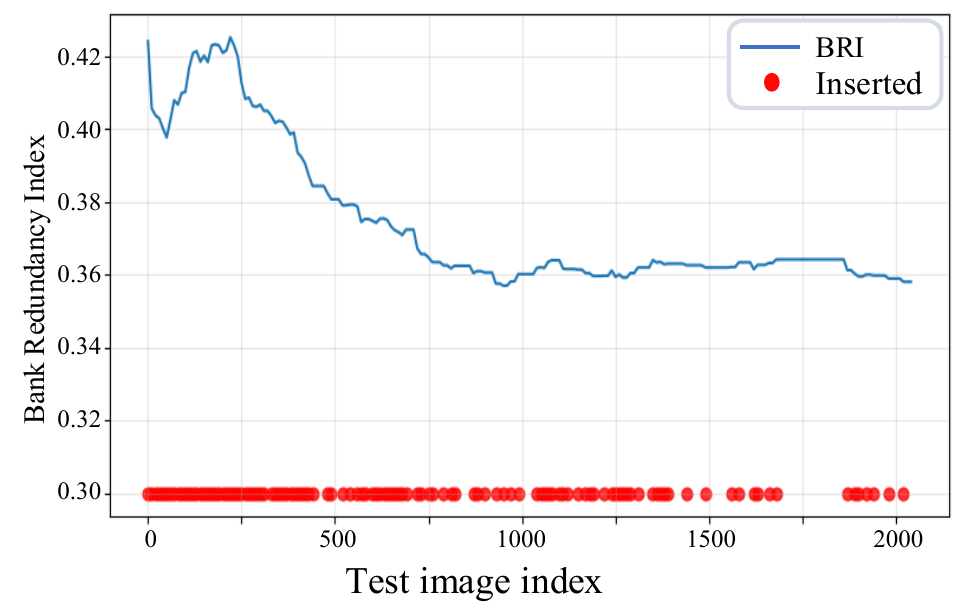}}
     \caption{ {Evolution of the feature bank during TTA. The blue line denotes the Bank Redundancy Index (BRI, lower = more diverse), and the red dots indicate test batches where new samples were inserted into the bank $\mathcal{F}$. }}
    \label{fig:bri_anchor_dice}
\end{figure}
\subsection{Ablation Analysis}

\subsubsection{Ablation Study of Each Component}
To assess the impact of each component within the A3-TTA framework, we conducted an ablation study on Domain B of the M\&MS dataset, and the results are presented in Table~\ref{tab:ablation}. The baseline (first row) utilized the source model without any adaptation. Next, we applied the mean-teacher model with cross-entropy loss ($\mathcal{L}_{mt}$), achieving a Dice score of 75.95\%. When anchor-target images filtered by entropy were introduced for semantic supervision ($\mathcal{L}_{sem}$), the Dice score was improved to 78.98\%, indicating that anchor-target images are crucial for enhancing adaptation. Adding the boundary-guided entropy minimization loss ($\mathcal{L}_{be}$) further boosted the performance to 79.38\%.
We then employed our proposed {CCD} to filter anchor-target images, leading to an average Dice of 80.36\%. Finally, the incorporation of Self-Adaptive EMA Rate (SER) for teacher updates raised the Dice score to 82.14\%.
These results demonstrate that each component contributes progressively to the overall performance improvement.

\subsubsection{Effect of Hyper-parameters}
The A3-TTA framework relies on three key hyper-parameters: the feature bank length ($L$) and the loss balance coefficients ($\beta$ and $\gamma$). To assess their sensitivity, we conducted experiments on domain B of the M\&MS dataset, evaluating various values of $L$, including 0, 5, 10, 20, 40, 80, and 160. Notably, setting $L=0$ means that $z_i$ is not replaced by $\hat{\boldsymbol{z}}_i$ to obtain $p'_i$, therefore bypassing the influence of both $\mathcal{L}_{sem}$ and $\mathcal{L}_{be}$. As illustrated in Fig.\ref{fig:hypa} (a), the performance stabilizes for $L \geq 5$, with $L=40$ striking the optimal balance between performance and computational efficiency. Similarly, Fig.\ref{fig:hypa}  (b) shows minimal fluctuation in the average Dice score across the range of $\beta$ values, with $\beta=5.0$ yielding the most robust results. Lastly, Fig.~\ref{fig:hypa} (c) reveals that $\gamma$ is more sensitive than the other two hyper-parameters, with $\gamma=1.0$ being the optimal setting for achieving the best balance between performance and stability.

\section{Analysis and Discussion}
\subsection{Analysis of the Anchor-Target Images}
First, to analyze how the ATIs and the feature alignment strategy works, we visualize the bottleneck features extracted by the model from source images, ATIs, and non-ATIs. In the T-SNE visualization, feature points positioned closer together indicate greater similarity.
As shown in Fig.~\ref{fig:tsne_2} (a), features from the source images cluster closely, while those from target images (ATIs and non-ATIs) are more dispersed. ATIs tend to align near the source image boundaries, whereas non-ATIs are further apart. This suggests that ATIs share similarities with the boundary cases of the source domain.
Additionally, we visualized the features of non-ATIs and ATIs after adaptation with our A3-TTA. As shown in Fig.~\ref{fig:tsne_2} (b), A3-TTA effectively aligns features by shifting non-ATI features closer to the ATI region, demonstrating the strong feature alignment capability of our method for non-ATIs.

Second, to assess the performance of our filtered ATIs on the source model and emphasize the superiority of the proposed CCD filtering method, we evaluated the Dice scores of ATIs filtered using entropy-based, Monte Carlo (MC) Dropout-based~\cite{gal2016dropout} filtering {(with 10 stochastic forward passes)}, and CCD methods. As shown in Fig.~\ref{fig:tsne_2}(c), under identical conditions and with the same number of selected slices, CCD-selected ATIs achieved {consistently and significantly} higher Dice than those obtained using entropy-based and MC Dropout-based filtering. Furthermore, the ATIs exhibited a higher average Dice score than the entire testing set, underscoring both the effectiveness of our ATIs and the efficiency of the CCD filtering method {(single forward pass vs. multiple forward passes)}.

{In addition, to show the effectiveness of our ATI bank update strategy in maintaining the diversity and dynamics of ATI images, we monitored the Bank Redundancy Index (BRI) that is the mean pairwise cosine similarity among features stored in the ATI bank $\mathcal{F}$. A lower BRI indicates more diverse samples in $\mathcal{F}$. As shown in Fig.~\ref{fig:bri_anchor_dice}, with the process of TTA, BRI  decreases gradually and then stabilizes at a relatively low level. In addition, the red dots in  Fig.~\ref{fig:bri_anchor_dice} show that the bank $\mathcal{F}$ is updated throughout the TTA process, which makes sure that $\mathcal{F}$ effectively tracks the distribution shift of the stream of testing data, indicating that an extra bank-cleaning mechanism is not needed in practice.}

\begin{table*}[!tb]
\centering
\caption{\label{continual_learning_ability_prostate} Evaluation of continual test-time adaptation performance and anti-forgetting ability using Dice scores for prostate MRI segmentation across multiple sequential target domains over multiple rounds. \dag denotes a significant improvement (p-value $<$ 0.05) over the best existing method.}
\begin{adjustbox}{width=1\linewidth,center=\linewidth}
\begin{tabular}{c|cccccc|c}
\hline \bigstrut[t]
Dice(\%) & \multicolumn{6}{c|}{Domain D $\xrightarrow{\hspace*{0.4cm}}$ Domain E $\xrightarrow{\hspace*{0.4cm}}$ Domain F $\xrightarrow{\hspace*{0.4cm}}$ Domain D $\xrightarrow{\hspace*{0.4cm}}$ Domain E $\xrightarrow{\hspace*{0.4cm}}$ Domain F} & Average $\uparrow$ \bigstrut[b]\\ 
\hline\bigstrut[t]
\multirow{1}{*}{Source Only} & 83.04±16.19 & 37.33±30.21 & 72.33±25.01 & 83.04±16.19 & 37.33±30.21 & 72.33±25.01 & 63.39±24.00 \\ 
\multirow{1}{*}{TENT~\cite{wang2021tent}}  & 84.53±14.91 & 71.06±25.16 & 79.35±20.14 & 83.65±16.02&67.54±28.74&75.50±24.12& 76.70±21.68   \\ 
\multirow{1}{*}{CoTTA~\cite{wang2022continual}}& 84.40±14.93 & 69.02±23.84 & 81.22±17.70 & 84.73±14.69 &76.18±20.34&82.89±15.58 &  79.51±17.98 \\ 
\multirow{1}{*}{SAR~\cite{niu2023towards}}& 84.22±15.05 & 63.03±25.25 & 78.37±20.33 & 84.07±15.32 &63.11±25.25&78.31±20.43 &  74.80±22.70 \\ 
\multirow{1}{*}{A3-TTA}  & \textbf{85.12±14.73} & \textbf{81.05±16.85} & \textbf{84.58±13.87} & \textbf{85.33±14.56}&\textbf{82.12±16.52}&\textbf{84.91±13.88}& \textbf{83.78±15.12}$^\dag$\bigstrut[b]\\  
\hline
\end{tabular}
\end{adjustbox}
\end{table*}

\subsection{Continual Learning and Anti-forgetting Ability}
In real clinical settings, the test target domain may shift over time, requiring the source model to sequentially adapt to multiple target domains. To  this end, we  evaluated the model's performance in a continual learning scenario on the prostate dataset, where the source model is first adapted to Domain D, then sequentially adapted to Domain E and Domain F. We compared A3-TTA  with the leading methods in Tables~\ref{tab:mms_sota_dice}, ~\ref{tab:mms_sota_assd} and~\ref{tab:prostate_assd_dice}, i.e., TENT~\cite{wang2021tent}, CoTTA~\cite{wang2022continual} and SAR~\cite{niu2023towards}.
As shown in the first three columns of Table~\ref{continual_learning_ability_prostate}, TENT improved the Dice score to 84.53\% on Domain D, followed by scores of 71.06\% on Domain E and 79.35\% on Domain F. CoTTA and SAR also demonstrated improvements during the first round of continual TTA. Our proposed A3-TTA achieved further improvements, with Dice scores of 85.12\%, 81.05\%, and 84.58\% on Domain D, E, and F, respectively, indicating the robustness of A3-TTA in continual learning scenarios.

To evaluate the anti-forgetting capability of A3-TTA, we conducted a second round of testing across these sequential target domains on the prostate dataset, and the results  are presented in columns 4-6 of Table~\ref{continual_learning_ability_prostate}. TENT exhibited a performance decline compared to the first round due to error accumulation over time, whereas CoTTA and SAR maintained relatively stable results. In contrast, A3-TTA consistently outperformed the existing methods in the second round. Notably, the average Dice score of A3-TTA in Round 2 surpassed its performance in Round 1 (84.91\% vs 84.58\% in Domain F). 
Furthermore, A3-TTA outperformed typical single-domain, single-epoch adaptation that obtained an average Dice score of only 81.07\% in Table~\ref{tab:prostate_assd_dice}.
These results demonstrate that A3-TTA not only maintains stability in continual TTA but also has a strong anti-forgetting ability. Furthermore, it benefits from continual learning, effectively leveraging knowledge from previous target domains to enhance performance over time.

\begin{table}[t]
  \centering
  \caption{{Robustness evaluation of different TTA methods on the M\&MS dataset under two types of noise. 
\dag denotes a significant improvement ($p<0.05$) over the best existing method. }}
  \begin{adjustbox}{width=0.5\textwidth}
  \begin{tabular}{l|l|ccc|c}
    \hline
    Noise & Method & Domain B & Domain C & Domain D & Average \bigstrut[t] \\
    \hline
    Clean & Source & 71.72±20.24 & 64.36±22.53 & 72.57±20.38 & 69.72 \\
    \hline
    \multirow{5}{*}{\makecell{Rician}}
    & Source & 50.83±30.02 & 42.40±28.61 & 42.89±28.67 & 46.72 \bigstrut[t]\\
    & TENT~\cite{wang2021tent} & 71.08±21.53 & 66.62±24.20 & 64.48±23.22 & 68.42 \\
    & CoTTA~\cite{wang2022continual} & 71.04±20.79 & 71.79±20.13 & 74.19±18.83 & 71.90 \\
    & SAR~\cite{niu2023towards} & 68.75±22.02 & 65.28±23.35 & 66.41±23.19 & 67.25 \\
        & A3-TTA
& \textbf{79.95±16.96}$^{\dag}$ & \textbf{77.24±18.81}$^{\dag}$ & \textbf{78.16±17.22}$^{\dag}$ & \textbf{78.78}$^{\dag}$ \bigstrut[b]\\
    \hline
    \multirow{5}{*}{\makecell{Motion \\ blur}}
    & Source & 60.51±20.10 & 44.21±23.82 & 66.51±19.33 & 56.93 \\
    & TENT~\cite{wang2021tent} & 67.69±20.87 & 65.20±22.03 & 64.21±21.71 & 66.24 \\
    & CoTTA~\cite{wang2022continual} & 69.16±19.82 & 69.89±19.09 & 72.69±18.98 & 70.10 \\
    & SAR~\cite{niu2023towards} & 64.81±20.75 & 63.12±20.80 & 64.84±21.80 & 64.32 \\
    & A3-TTA & \textbf{75.44±16.75}$^{\dag}$ & \textbf{74.21±16.75}$^{\dag}$ & \textbf{74.98±17.50}$^{\dag}$ & \textbf{74.98}$^{\dag}$ \\
    \hline
  \end{tabular}
  \end{adjustbox}
  \label{tab:mms_noise}
\end{table}
\subsection{{Robustness to Noise}}
{To assess the robustness of A3-TTA to acquisition noise, we experimented  on the M\&MS dataset with two clinically motivated perturbations: 1) Rician noise ($\sigma$=0.05) that simulates magnitude-domain acquisition noise in MRI, and 2) motion blur ($k$=12) that simulates in-plane patient motion during scanning. 
As reported in Table~\ref{tab:mms_noise}, the source model has a large Drop of average Dice from 69.72\% to 46.72\% under the Rician noise, and to  56.93\% under motion blur. In contrast, our A3-TTA obtained an average Dice of 78.78\% under the Rician noise, which significantly outperformed the compared methods, with only a small performance drop compared with clean testing data (80.12\% in Table~\ref{tab:mms_sota_dice}). 
For the motion blue situation, our A3-TTA also maintained a significantly higher Dice (74.98\%) compared with the best existing method (CoTTA~\cite{wang2022continual} with Dice of 70.10\%) and the source only model (56.93\%). 
The results indicate A3-TTA's stronger resistance to device noise and motion artifacts than the other TTA methods. }

\subsection{Computational Analysis}
{As our method does not introduce new parameters or architecture changes to the source model, the model size after adaptation is identical to that of the source model.} 
{The practical difference  from existing TTA methods lies in the subset of {trainable} parameters during adaptation: prompt-based methods (VPTTA~\cite{chen2024each}, EDCP~\cite{liu2025efficient}) optimize only lightweight prompts (tens to hundreds of parameters), BN\textendash only approaches (TENT~\cite{wang2021tent}, SAR~\cite{niu2023towards}, GraTa~\cite{chen2025gradient}) update only BN affine parameters ($\approx2.9K$), whereas  MT~\cite{tarvainen2017mean}, CoTTA~\cite{wang2022continual}, and our A3\textendash TTA update all parameters ($\approx1.81M$ for 2D U-Net).} {Despite the increase of trainable parameters, it does not increase model size,  and typically delivers stronger adaptation performance.}

For computational analysis, we compared our method with CoTTA~\cite{wang2022continual}, which achieved the best performance among existing methods reported in Table~\ref{tab:mms_sota_dice}. {Under the same backbone and setting,} with a feature bank size of $L = 40$ and a batch size of 10, the slice\textendash level adaptation time on the M\&MS dataset is 0.044 seconds, using 4.99 GB of GPU memory. {Both our method and CoTTA update all backbone parameters, yet our method is faster and more memory-efficient:} 
CoTTA takes 0.060 seconds and consumes 7.23 GB of GPU memory under the same configuration.

\subsection{Limitations {and Future Works}}
{Our method targets TTA for image segmentation and is effective under typical distribution shifts in both {medical} settings (variations in imaging devices, hospitals, and patient demographics) and {real-world} scenes (different weather conditions). A3-TTA establishes anchor–target images within the target domain to facilitate the adaptation of subsequent test images. However, when domain shifts are extreme, particularly cross-modality differences such as CT$\leftrightarrow$MRI, the target-domain anchors may deviate substantially from the source distribution, potentially limiting overall adaptation performance.}

{Another limitation concerns the network architecture and computational setup. Due to GPU memory constraints, we conducted all experiments with a 2D backbone rather than a 3D segmentation network. While this choice is consistent with prior TTA studies on these datasets and mitigates issues from large inter-slice spacing, it may reduce access to full volumetric context and necessitates a sliding-window strategy for 3D volumes. Furthermore, the feature bank in A3-TTA is initialized by the first few samples. This design ensures fairness under the strict TTA setting, where neither source nor target data are available beforehand, but it may cause slightly unstable adaptation during the initial iterations.}  

{Looking forward, several research directions are worth pursuing. First, given that the quality of feature bank $\mathcal{F}$ is relatively low at the beginning, it is worthwhile to improve the adaptation performance for the first few samples with some complementary strategies such as lightweight self-supervised objectives or entropy minimization without violating the TTA setting. Second, extending anchor-guided adaptation to handle larger cross-modality gaps such as CT$\leftrightarrow$MRI remains an important challenge to address. In addition, integrating anchor-target images with emerging foundation models~\cite{IPLC2025} could allow anchors to act as strong reference prompts for obtaining high-quality pseudo-labels. } 

\section{Conclusion}
We presented A3-TTA, an anchor-guided test-time adaptation framework for {image segmentation} under domain shift. 
It first identifies relatively well-predicted target domain  samples as Anchor-Target Images (ATIs) via a class compact density metric. The latent features of ATIs are stored in a feature bank, which serves as an intermediate reference to refine the predictions of other test samples. The resulting pseudo labels are further regularized via semantic consistency and boundary-aware entropy minimization.
To ensure stable model updates under single-pass test-time constraints, A3-TTA integrates a mean-teacher framework with a self-adaptive EMA strategy, dynamically adjusting the update rate based on prediction consistency. Extensive experiments on {three} public multi-domain datasets demonstrated the superiority of A3-TTA over state-of-the-art TTA approaches in both static and continual adaptation settings, and it exhibited strong anti-forgetting ability in continual learning.
In the future, it would be of interest to extend A3-TTA to 3D segmentation models and explore its applicability to more challenging scenarios, such as cross-modality and extreme domain shifts.

\bibliographystyle{IEEEtran}
\bibliography{myref}

\vfill

\end{document}